\documentclass{article}

\usepackage{graphicx}
\usepackage{subcaption}
\usepackage{wrapfig}
\usepackage{caption}

\usepackage[nonatbib,preprint]{nips_2018}

\usepackage[utf8]{inputenc} 
\usepackage[T1]{fontenc}    
\usepackage{hyperref}       
\usepackage{url}            
\usepackage{booktabs}       
\usepackage{amsfonts}       
\usepackage{nicefrac}       
\usepackage{microtype}      
\usepackage{amsmath,amssymb}
\usepackage{graphicx}
\usepackage{color}
\usepackage{placeins}

\title{Learning to Collaborate for User-Controlled Privacy}

\author{
  Martin Bertran \textsuperscript{1\dag} \\
  \texttt{martin.bertran@duke.edu} \\
  \And
  Natalia Martinez \textsuperscript{1\dag*} \\
  \texttt{natalia.martinez@duke.edu} \\
  \And
  Afroditi Papadaki \textsuperscript{2}\\
  \texttt{a.papadaki.17@ucl.ac.uk} \\
  \And
  Qiang Qiu \textsuperscript{1}\\
  \texttt{qiuqiang@gmail.com} \\
  \And
  Miguel Rodrigues \textsuperscript{2}\\
  \texttt{m.rodrigues@ucl.ac.uk} \\
  \And
  Guillermo Sapiro \textsuperscript{1}\\
  \texttt{guillermo.sapiro@duke.edu}
}

\begin{document}
\maketitle
{\begin{center}
\textbf{1} Duke University, Durham, NC, USA;\\
\textbf{2} University College London, London, UK;\\
\textbf{*} Corresponding author\\
\dag These authors contributed equally to this work.
\end{center}
}
\vspace{0.2in}

\noindent
\textit{Privacy is a human right.} Tim Cook, Apple CEO.

\begin{abstract}

It is becoming increasingly clear that users should own and control their data. Utility providers are also becoming more interested in guaranteeing data privacy. As such, users and utility providers should \textit{collaborate} in data privacy, a paradigm that has not yet been developed in the privacy research community. We introduce this concept and present explicit architectures where the user controls what characteristics of the data she/he wants to share and what she/he wants to keep private. 
This is achieved by collaborative learning a sanititization function — either a deterministic or a stochastic one — that retains valuable information for the utility tasks but it also eliminates necessary information for the privacy ones.
As illustration examples, we implement them using a plug-and-play approach, where no algorithm is changed at the system provider end, and an adversarial approach, where minor re-training of the privacy inferring engine is allowed. In both cases the learned sanitization function keeps the data in the original domain, thereby allowing the system to use the same algorithms it was using before for both original and privatized data. We show how we can maintain utility while fully protecting private information if the user chooses to do so, even when the first is harder than the second, as in the case here illustrated of identity detection while hiding gender. 
\end{abstract}

\section{Introduction, Challenges, and Contributions}
\label{sec:introduction}

Advances in machine learning allow to develop products that leverage individuals data to provide valuable services to them. At the same time care must be taken to provide an appropriate protection for users in order to adhere to various privacy, legal, and ethical constraints. Critical to this is to develop a trusting-based system where the data receiver can infer from shared data information a user does not consider private but concurrently cannot infer from the data information the user desires to keep private. Each user should have the ability to define sensitive and non-sensitive information associated with her/his data, which may differ from user to user; we propose that a user and the service provider \textit{collaborate} towards achieving \textit{user-specific} privacy.
 
We address the following scenario: An entity (e.g., a bank) wants to offer a service to its users based on their data (e.g., an ATM verify the card holder is the legitimate owner). However, some users want to be able to prevent the entity from inferring certain information from their data  (e.g., gender), whereas other users may wish to prevent the entity from inferring other attributes (e.g., ethnicity). We assume the entity and each individual user \textit{collaborate} in creating such system; it is as important for the user to preserve her/his privacy as it is to the entity to guarantee it, being this a mutually beneficial system. First, the users will be more comfortable because their privacy is respected. Second, the entity offering the service increases user's trust and provides a principled approach to guarantee that they are not using/inferring users' sensitive information in case of legal/ethical disputes (e.g., their system can be audited to show the private information cannot be extracted). 

This proposed new paradigm of collaborative privacy environment is critical since it has been shown once and again that, e.g., algorithmic or data augmentation and unpredictable correlations, can easily break privacy \cite{Israel2014,Narayanan2008,Oh2016,Reuben2016}. The impossibility of universal privacy protection has also been studied extensively in the domain of differential privacy \cite{Dwork2008}, where a number of authors have shown that assumptions about the data or the adversary must be made in order to be able to provide utility  \cite{Dwork2010, Hardt2016, Kifer2011a,  Kifer2014}.\footnote{Privacy is not hacking, the first one is the challenge addressed in this work, the second is a security/encryption task. Both are closely connected, and removing private data will immediately improve security.}

Therefore, given each individual user specific privacy requirements, it is important to design collaborative systems where each individual user shares a sanitized version of their data with the service provider in such a way that user-defined non-sensitive tasks can be performed but user-defined sensitive ones cannot, with the set of machine learning algorithms available at the service provider. A sanitization mechanism 
should not alter the data format so that the service provider machine learning system, consisting of a set of algorithms to infer user attributes from their data, can work independently of the user privacy preferences.

\textbf{Contributions-}
We propose a novel framework that assumes a \textit{system}-and-\textit{user} \textit{collaborative} environment in order to allow a service provider to infer certain user attributes but not others depending on \textit{user-specific} privacy preferences. The proposed framework is based on the use of user-specific privacy filters that retain important information for the entity to perform legitimate tasks but simultaneously filter out information necessary for the entity to perform sensitive non-legitimate ones. The proposed framework is agnostic to the presence or absence of such user-specific privacy filters during operation, since by design it can handle both the original and the filtered/privatized data. {The service provider uses a single processing pipeline on both sanitized and original data and still guarantee to each individual user that their privacy will be preserved, while still providing the required utility service.}

Our proposed framework is also flexible to allow different levels of collaboration between each user and the service provider, from re-training of all key components (utility algorithm, private data inferring algorithm, sanitization function) to re-training only some components, or to re-train only the user-specific privacy filters for 'plug-and-play' operation. Some of these variations are exemplified here both via deterministic and stochastic sanitization mechanisms that show how we can, without affecting utility, practically achieve the (optimal privacy) prior distribution for the privacy.

The proposed framework is described in Section ~\ref{sec:model}, and learning architectures for it are given in
Section \ref{sec:architecture}.
Examples are presented in Section~\ref{sec:results}. 
{Since data should be protected for privacy before being transmitted (even at the sensor level if needed), we present results on a prototype hardware platform (to be demonstrated at the conference) that implements the framework at the data sharing step.}
We discuss connections with the literature in Section \ref{sec:literature}.
The paper is concluded in Section~\ref{sec:conclusion}.

\section{The Proposed Collaborative Privacy Model}
\label{sec:model}

\begin{wrapfigure}{L}{0.5\textwidth}
\centering
\includegraphics[width=0.4\columnwidth]{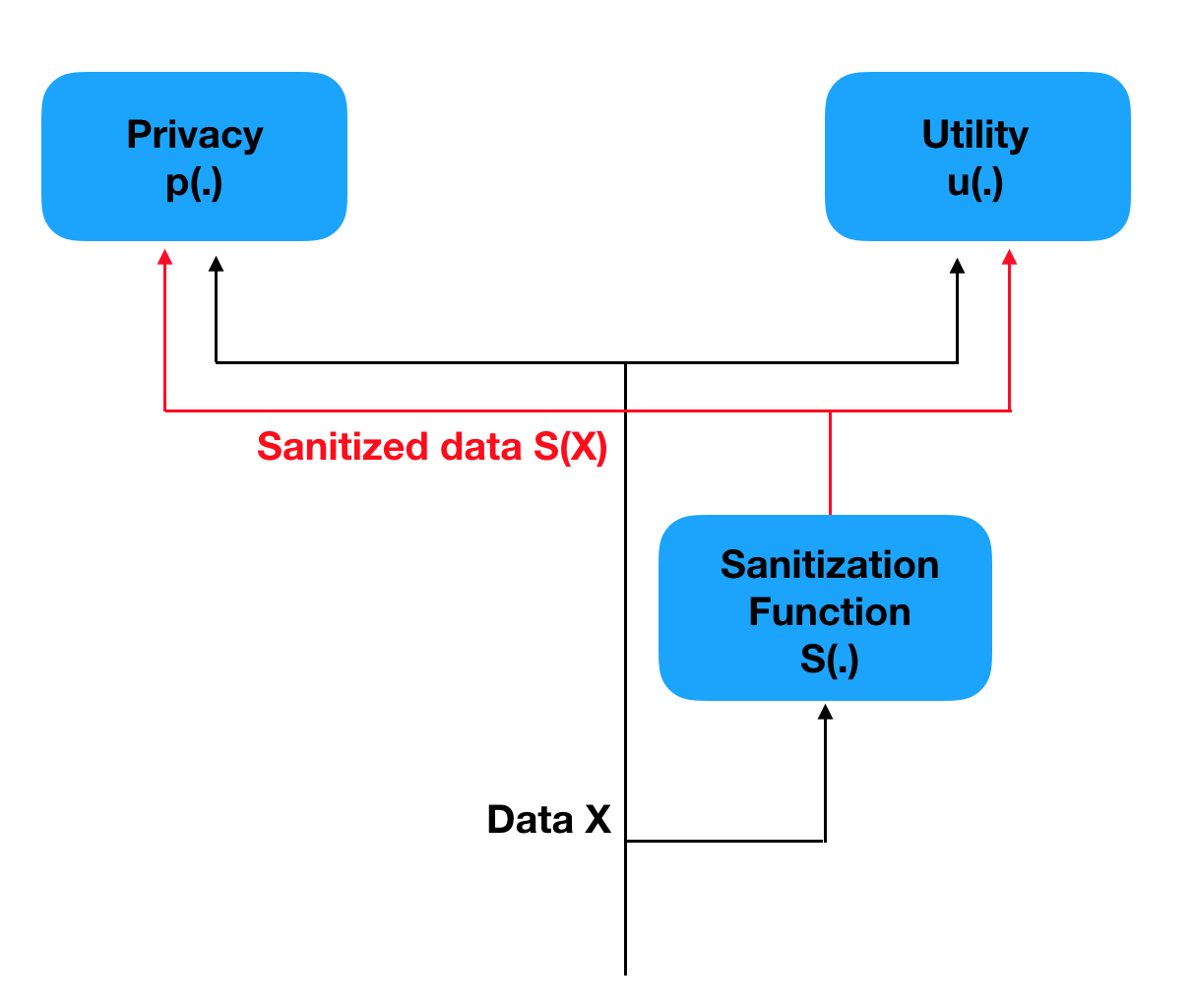}
\caption{\it Three components of the collaborative privacy framework {from the perspective of a user per her/his privacy preferences}. Raw data can be directly fed into the privacy and utility tasks. Since the sanitization function is a transformation that maps a space onto itself, the sanitized data can also be directly fed to both tasks without any need for further adaptations.}
\label{fig:model}
\end{wrapfigure}

The proposed collaborative privacy framework is based on various desiderata:
\textbf{(1)} The entity must provide the user concrete guarantees. In particular, given a set of algorithms used by the entity to infer possible user attributes, the entity should be able to guarantee the user what it can and what it cannot infer given the data conveyed by the user to the entity; \textbf{(2)}
The user must be able to define what she/he would like the entity to be able to infer (utility); \textbf{(3)} The user should be able to define what she/he would like the entity not to be able to infer (privacy); and \textbf{(4)} The approach should integrate seamlessly (i.e., 'plug-in') within current systems, in the sense that the entity will not have to alter its existent state-of-the-art inference algorithms in order to allow for user-specific non-sensitive tasks and block user-specific sensitive ones. This is specially important in scenarios where an entity may use many different state-of-the-art algorithms to infer all kinds of user attributes from user data, particularly in platforms with a very large number of users with very different privacy requirements.
    
Following this, {from the perspective of a specific user}, the system has three key components, Figure \ref{fig:model}, \textbf{(1)} The \textit{utility} component is the service providing function; it takes input data, either raw or sanitized data, and infers a variable of interest;
\textbf{(2)} The \textit{privacy} component also operates on the same input data (raw or sanitized, they share the same format), and attempts to infer sensitive information from it; and \textbf{(3)}
 The \textit{sanitization} function is a learned transformation that maps the input data onto the same space (i.e., maps images to images with the same dimensions).
Note that the sanitization function -- mapping user data from the input space to the same space -- is critical: it allows the user to share ``filtered'' data with the entity without requiring (potentially) any alteration to the state-of-the-art (utility) inference algorithms.

It is not immediately clear if it is possible to define sanitization functions that transform the data in such a way that the set of pre-trained algorithms for the legitimate tasks perform well while the set of pre-trained algorithms for the sensitive tasks under-performs when the user desires that. For example, for visual data such sanitization functions have to transform an image onto another ``image'' where information relevant for the utility tasks is preserved whereas information associated with the sensitive tasks is blocked. No current approaches -- as reviewed in later sections -- have this property.
Given the utility components and privacy components (here a single utility and privacy component are considered, but the principles generalize to systems with multiple such components), our goal is to learn a sanitization function that preserves performance on the utility task and lowers performance on the privacy task.

Let $D_{KL}(p\mid\mid q)$ be the Kullback-Leibler divergence from probability distribution $q$ to probability distribution $p$; $P(u\mid x)$ the posterior probability distribution of the utility variable $u$ conditioned on the raw data $x$; $P(u\mid S(x))$ the posterior probability distribution of the utility variable $u$ conditioned on the sanitized data $S(x)$;
$P(p \mid S(x))$ the posterior probability distribution of the privacy variable $p$ conditioned on the sanitized data $S(x)$; and $P(p)$ the prior distribution of the privacy variable.
We propose the following loss to train the sanitization function:
\begin{equation}
    \label{eq:sanitizationLoss}
    \begin{split}
    \mathrm{Loss}_{S} &= (1-\alpha) D_{KL}(P(u\mid x)\mid\mid P(u\mid S(x))) + \alpha  D_{KL}(P(p)\mid\mid P(p\mid S(x))).
    \end{split}
\end{equation}

The proposed loss function is such that it attempts to guarantee that the outputs of the utility algorithm given the true ($x$) or sanitized ($S(x)$) data are indistinguishable, and concurrently that the output of the privacy algorithm given the sanitized data is indistinguishable from the privacy variable prior. The parameter $\alpha \in (0,1)$ controls the trade-off between preserving information on the utility variable, and destroying information on the privacy variable.
{
We further emphasize the importance of the mapping constraint on the sanitization function. By preserving the original space of the input data, the sanitization function can work \emph{in tandem} with both the privacy and utility algorithms without any need to modify existing architectures.
It is of course not immediately clear that it is possible to simultaneously filter out a privacy variable while maintaining performance on the utility variable in the raw input space. One of the contributions of this work is to show
validations of this concept.

There are several possible training approaches we can pursue, some of them exemplified in this work. We can choose to exclusively train the sanitization function using the proposed loss, leaving the existing privacy and utility algorithms untouched. This training approach is very appropriate for the simple plug-in scenario, since no modification of the pre-trained utility and privacy algorithms is needed.

We can also \emph{adversarially} train the privacy task. In this scenario, the privacy task is attempting to defeat the sanitization function and simultaneously perform well on unsanitized data. The sanitization function is attempting to fool the privacy task (approximate the privacy prior) and simultaneously maintain performance on the utility task. This training is computationally more expensive, but opens the door to more universal privacy guarantees, being adapted to the architecture but not to its specific parameters (weights); see captions in Figure \ref{fig:Dkl}.

Finally, the utility function can be trained \emph{collaboratively} with the sanitization function; the utility function is aware of the sanitized data and can collaborate with the sanitization function to better preserve utility performance.

Naturally, these approaches can be freely combined to best suit the user’s and entity’s needs.
In the following section, we exemplify the plug-and-play and adversarial training approaches.}

\section{Privacy Learning Architectures}
\label{sec:architecture}

\begin{wrapfigure}{L}{0.5\textwidth}
    \centering
    \begin{subfigure}[b]{0.23\textwidth}
        \includegraphics[width=\textwidth]{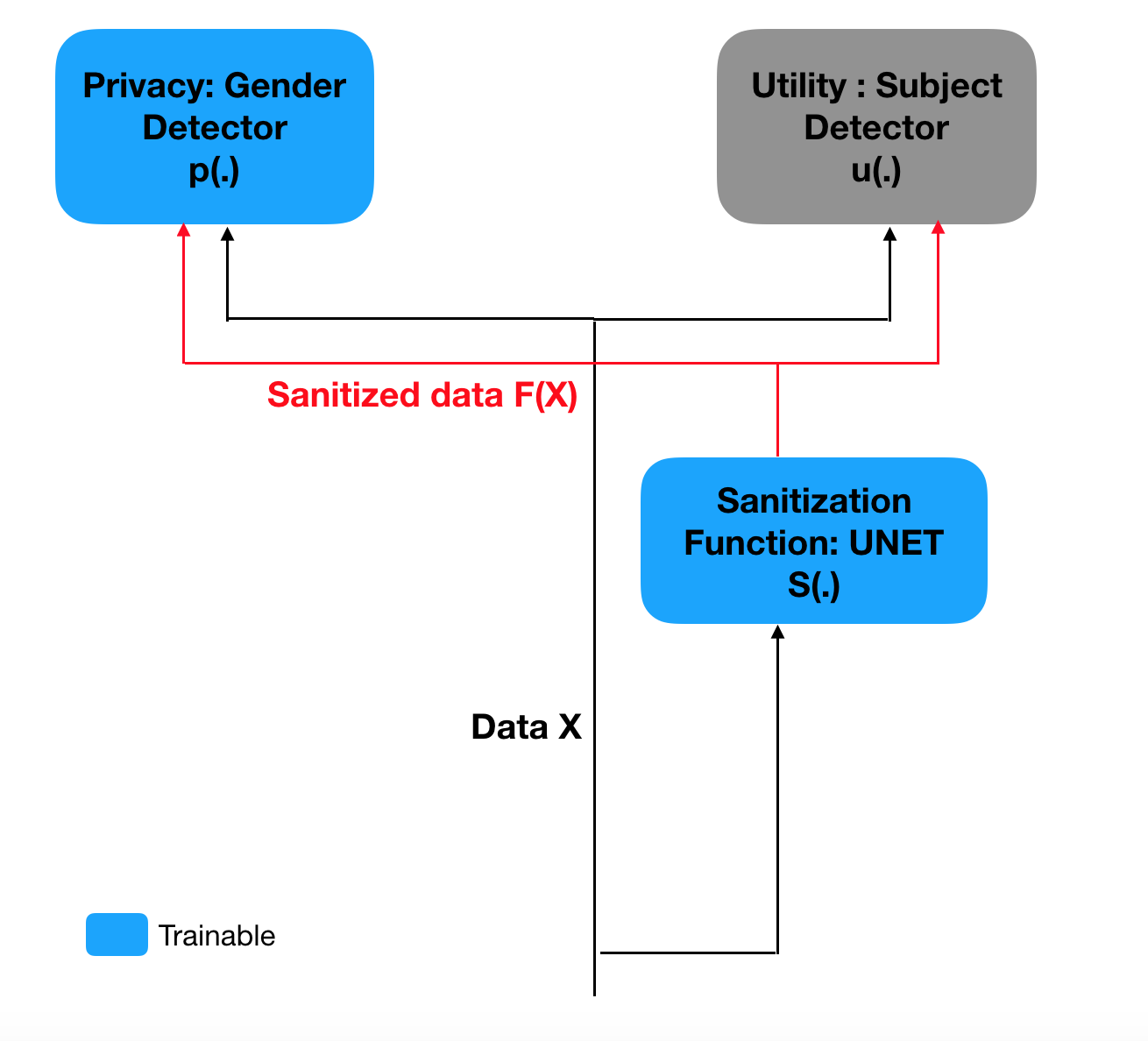}
        \caption{\it Deterministic}
        
    \end{subfigure}
    ~ 
    \begin{subfigure}[b]{0.25\textwidth}
        \includegraphics[width=\textwidth]{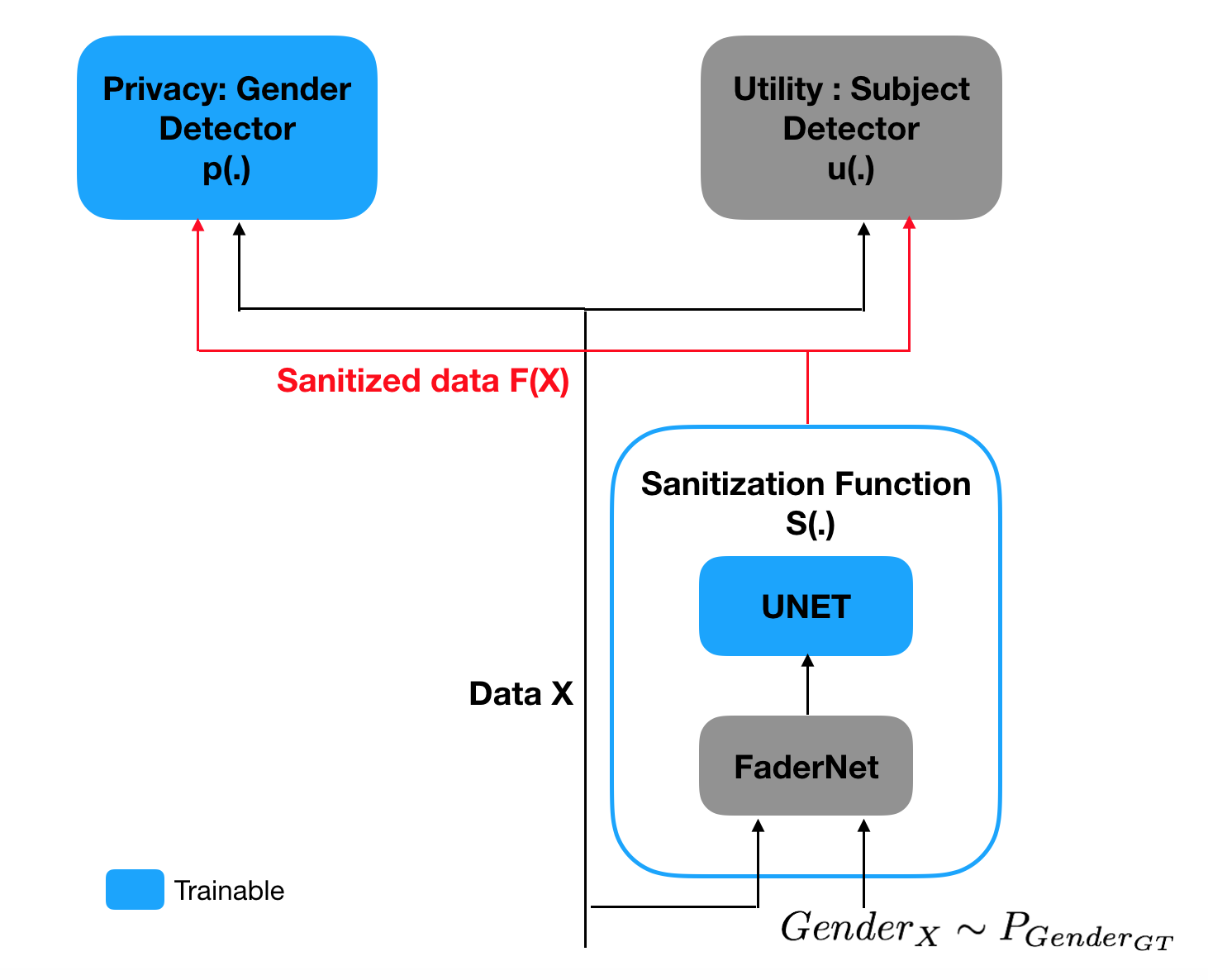}
        \caption{\it Stochastic}
    \end{subfigure}
    \caption{\it Proposed collaborative privacy learning architectures/models. In the stochastic model the raw image is fed through the gender FaderNet function, where its original gender is overwritten at random. Here the gender attribute is drawn from the prior gender distribution of the dataset. This re-sampled image is then fed into a fully-trainable UNET.
    In the deterministic model the data is fed directly to the fully-trainable UNET. In both cases the sanitized and raw data are inputs to the privacy and the utility tasks. The UNET and the privacy blocks are trainable, while the weights of the utility model and the FaderNet remain fixed.}
\label{fig:models}
\end{wrapfigure}

To illustrate the proposed framework, we chose subject recognition (harder task) and gender recognition (easier task) as our utility and privacy respectively. 
We propose two different sanitization architectures, a \textit{stochastic} one and a \textit{deterministic} one, Fig ~\ref{fig:models}. We implement them using the plug-and-play and the adversarial approaches mentioned above. In the deterministic architecture, the data is fed directly into a fully-trainable UNET, \cite{ronneberger2015u}, whose objective is to fulfill the sanitization task. In the stochastic approach the sanitization function first randomly overwrites the gender attribute of the input image using the pretrained FaderNet \cite{FaderNetworks}, done by sampling the gender of each of the images from the prior gender distribution of the dataset, and then feeds the resulting image through a fully-trainable UNET.

The pretrained FaderNet was chosen as the base for the stochastic sanitization function since this network is already trained to defeat a gender discriminator in its \textit{encoding space} (privacy variable is concealed here). This proves a suitable baseline comparison and starting point for a sanitization function that needs to fool a gender discriminator \emph{in image space} while simultaneously preserving subject recognition performance. We show below the performance of using only the pretrained gender FaderNet and demonstrate how training a UNET on top of it improves both the utility and privatization performances.

For both architectures, the UNET portion of the sanitization function was trained using the proposed loss function in Eq. (\ref{eq:sanitizationLoss}). For the adversarially trained results, the final layer of the privacy function (DEXNet) is trained using the  binary cross entropy (BCE) loss function,
\begin{equation}
    \label{eq:privatizationLoss}
    \begin{split}
    \mathrm{Loss}_{P} &=  BCE(y_p(x), p(x)) +  BCE(y_p(x), p(S(x))),
    \end{split}
\end{equation}
\noindent where $y_p(x)$ is a one-hot encoding of the ground truth gender label. This adversarial loss equally weights the performance of the gender detector on both the raw (unsanitized) data  ($x$) and the sanitized data ($S(x)$). The sanitization and privacy tasks are playing an adversarial game. Adversarial training provides stronger privacy guarantees since the privacy detector is unable to infer the privacy variable even after retraining.
We alternate the training of the DEXNet and UNET every one epoch, in the supplementary methods section we show how the losses $Loss_S$ and $Loss_P$ evolve in each iteration. 
 
\section{Experimental Results}
\label{sec:results}

Our data consisted of 15k samples (10k train, 5k test) from the realigned FaceScrub dataset \cite{Ng2014}.
The identification utility task is performed using the ResNet-50 implementation of VGGFace2 \cite{cao2017vggface2}, where the final dense layer was retrained on the chosen dataset. For gender recognition (privacy) we used a pretrained WideResNet-based, \cite{zagoruyko2016wide}, implementation of DEXNet \cite{rothe2018deep}.

\begin{figure}[!ht]
    \centering
    
    \begin{tabular}{cc}
    \begin{subfigure}[b]{0.4\textwidth}
        \includegraphics[width=\textwidth]{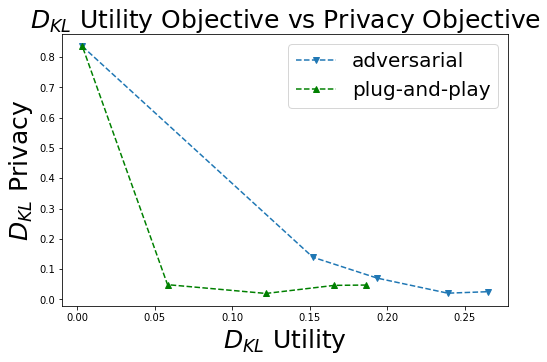}
        \caption{\it Deterministic model}
    \end{subfigure}&
    
        \begin{subfigure}[b]{0.4\textwidth}
        \includegraphics[width=\textwidth]{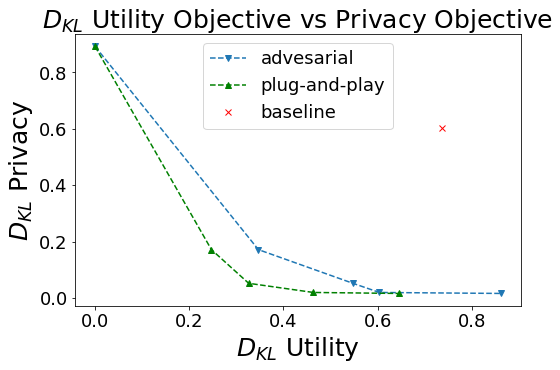}
        \caption{\it Stochastic model}
    \end{subfigure}
    
    ~ 
      \end{tabular}
      
    \caption{Trade-off achieved by the sanitization function between the KL-divergence of the utility distribution from the objective ($D_{KL}(P(u|x)||P(u|S(x)))$) and the KL-divergence of the gender distribution from the prior ($D_{KL}(P(p)||P(p|S(x)))$). Compared to the adversarial approach, we observe better performance of the tailored plug-and-play (task is inherently easier). Thereby it is important to show the still excellent performance of the ``more universal'' adversarial training (tested gender networks that performed the same for unfiltered data and adversarial filtered data, the privacy $D_{KL}$ deteriorated by up to 90\% on the plug-and-play sanitization).
    For $\alpha = 0$ no sanitization occurs and there is no loss in utility performance. As $\alpha$ increases ($\alpha=0, 0.05, 0.2, 0.5, 0.8$ in the figure), the KL-divergence for the utility increases and the privacy performance quickly approaches the prior.}
    \label{fig:Dkl}
    
\end{figure}

Figure \ref{fig:Dkl} shows the utility/privacy trade-off achieved by the stochastic and deterministic sanitization functions over the tested $\alpha$ values, as measured by the KL divergence for both the plug-and-play and adversarial training. As $\alpha$ increases, the privacy task quickly approaches the prior distribution (KL divergence goes to zero), while the utility task correspondingly loses performance (KL divergence increases). We also observe that the baseline performance of the pretrained FaderNet comes at a considerable cost in utility performance. We also observe that the full stochastic architecture trained following Eq. (\ref{eq:sanitizationLoss}) is able to simultaneously recover performance on the utility task and improve performance on the privacy task. For this particular example, the sanitization function trained with Eq. (\ref{eq:sanitizationLoss}) shows a net improvement over both the privacy and utility tasks with no downside.

\begin{figure}[!ht]
    \centering
    
    \begin{tabular}{cc}
    \begin{subfigure}[b]{0.4\textwidth}
        \includegraphics[width=\textwidth]{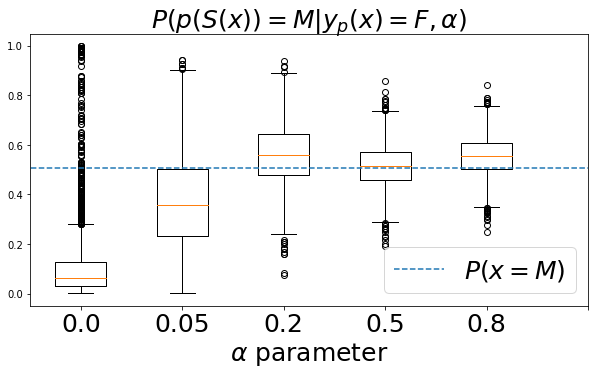}
        \caption{\it $P(p(S(x)) = M | y_p(x) = F, \alpha)$, Deterministic model}
    \end{subfigure}&
    
        \begin{subfigure}[b]{0.4\textwidth}
        \includegraphics[width=\textwidth]{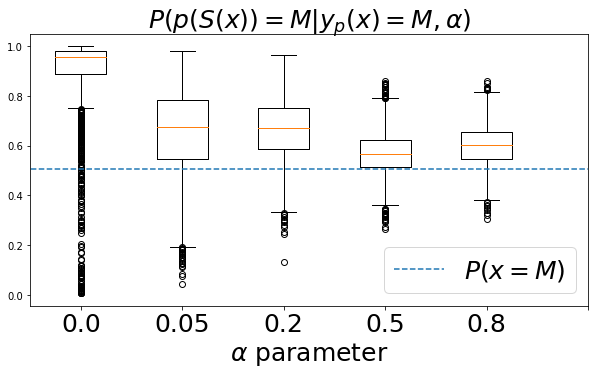}
        \caption{\it $P(p(S(x)) = M | y_p(x) = M, \alpha)$, Deterministic model}
    \end{subfigure}\\
    
    ~ 

    \begin{subfigure}[b]{0.4\textwidth}
        \includegraphics[width=\textwidth]{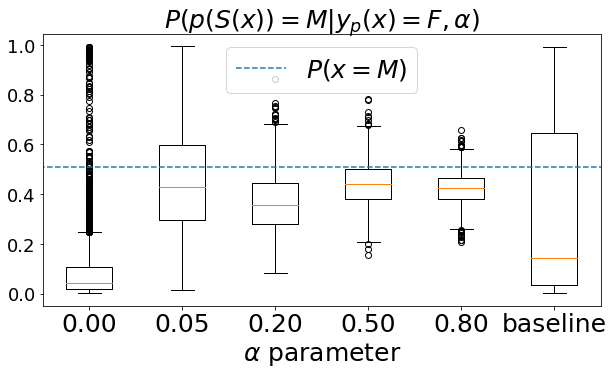}
        \caption{\it $P(p(S(x)) = M | y_p(x) = F, \alpha)$, Stochastic model}
    \end{subfigure}&
    
        \begin{subfigure}[b]{0.4\textwidth}
        \includegraphics[width=\textwidth]{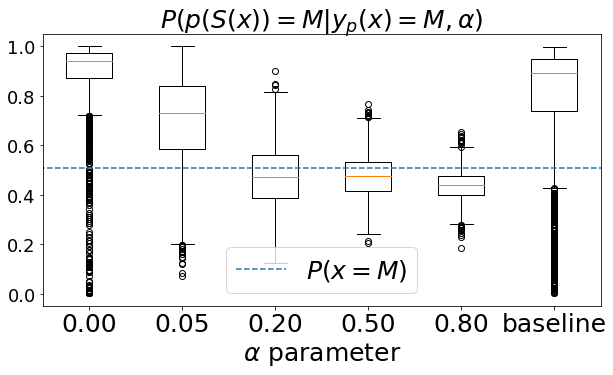}
        \caption{\it $P(p(S(x)) = M | y_p(x) = M, \alpha)$, Stochastic model}
    \end{subfigure}
    
    ~ 
      \end{tabular}

    \caption{\it Detailed breakdown of the gender probabilities computed by the privacy task as a function of $\alpha$ for the test dataset; training was done adversarially. Whisker plots show median and interquartile ranges, with outliers shown as circles.
    The first column shows the output probability of being male (M) as computed by the gender discriminator over all female (F) subjects. Second column shows the same output probabilities for all male subjects. Results are shown as a function of $\alpha$, where $\alpha=0$ is the performance over raw data. The dotted blue line indicates the prior probability of the dataset.}
\vspace{-.2in}
    \label{fig:pgenderAdv}
\end{figure}

\begin{figure}[!ht]
    \centering
    
    \begin{tabular}{cc}
    \begin{subfigure}[b]{0.4\textwidth}
        \includegraphics[width=\textwidth]{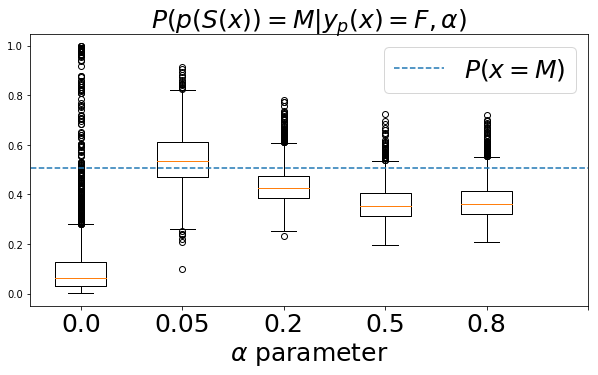}
        \caption{\it $P(p(S(x)) = M | y_p(x) = F, \alpha)$, Deterministic model}
    \end{subfigure}&
    
        \begin{subfigure}[b]{0.4\textwidth}
        \includegraphics[width=\textwidth]{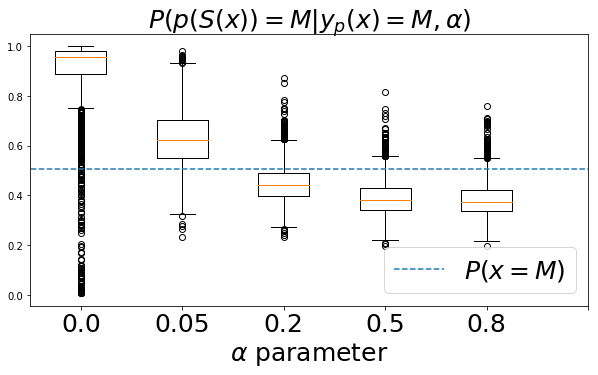}
        \caption{\it $P(p(S(x)) = M | y_p(x) = M, \alpha)$, Deterministic model}
    \end{subfigure}\\
    
    ~ 

    \begin{subfigure}[b]{0.4\textwidth}
        \includegraphics[width=\textwidth]{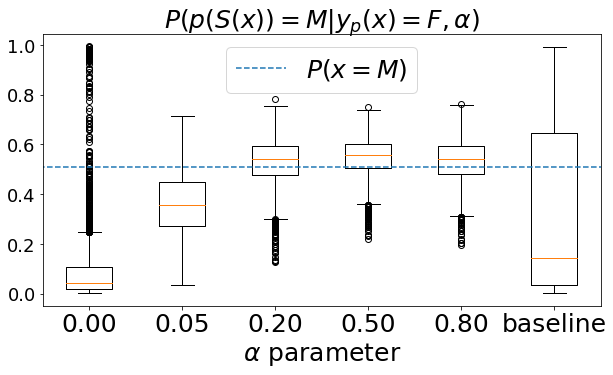}
        \caption{\it $P(p(S(x)) = M | y_p(x) = F, \alpha)$, Stochastic model}
    \end{subfigure}&
    
        \begin{subfigure}[b]{0.4\textwidth}
        \includegraphics[width=\textwidth]{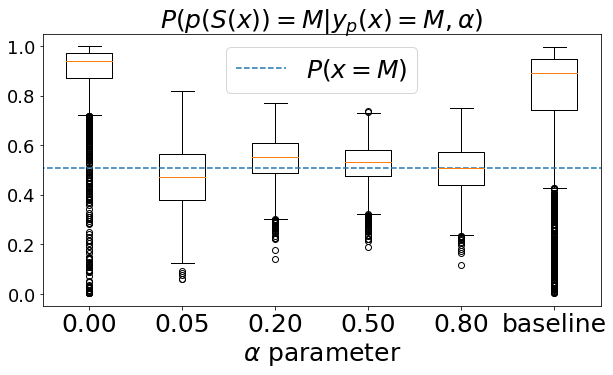}
        \caption{\it $P(p(S(x)) = M | y_p(x) = M, \alpha)$, Stochastic model}
    \end{subfigure}
    
    ~ 
      \end{tabular}

    \caption{\it Detailed breakdown of the gender probabilities computed by the privacy task as a function of $\alpha$, training was done in a plug-and-play approach. Whisker plots show median and interquartile ranges, with outliers shown as cricles.
    The first column shows the output probability of being male (M) as computed by the gender discriminator over all female (F) subjects. Second column shows the same output probabilities for all male subjects. Results are shown as a function of $\alpha$, where $\alpha=0$ is the performance over raw data. The dotted blue line indicates the prior probability of the dataset.}
    \vspace{-.1in}
    \label{fig:pgenderPAP}
\end{figure}

\begin{figure}[!ht]
    \centering
    
    \begin{tabular}{cc}
    \begin{subfigure}[b]{0.4\textwidth}
        \includegraphics[width=\textwidth]{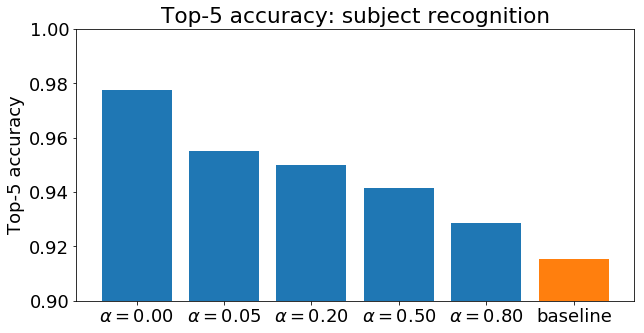}
        \caption{\it Top-5 Categorical accuracy, Deterministic model, adversarial approach}
    \end{subfigure}&
    
        \begin{subfigure}[b]{0.4\textwidth}
        \includegraphics[width=\textwidth]{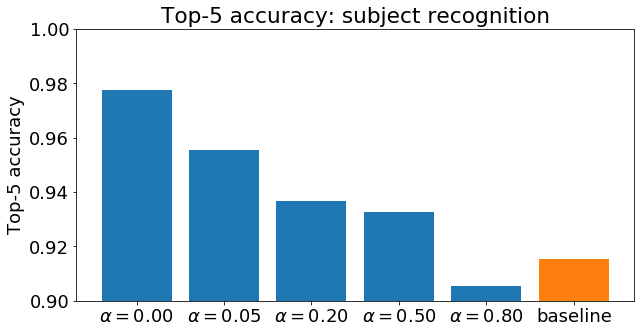}
        \caption{\it Top-5 Categorical accuracy, stochastic model, adversarial approach}
    \end{subfigure}\\
        \begin{subfigure}[b]{0.4\textwidth}
    \includegraphics[width=\textwidth]{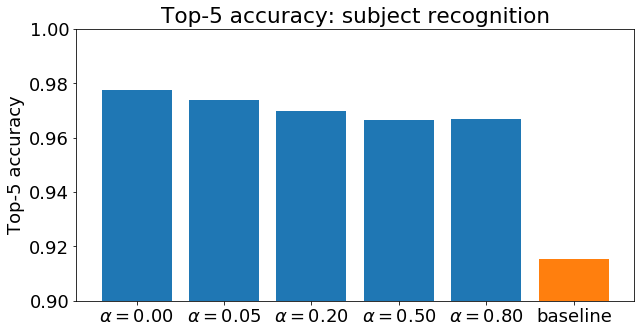}
        \caption{\it Top-5 Categorical accuracy, Deterministic model, plug-and-play approach}
    \end{subfigure}&
    
        \begin{subfigure}[b]{0.4\textwidth}
        \includegraphics[width=\textwidth]{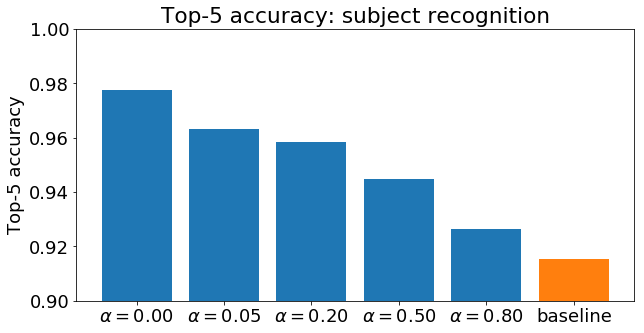}
        \caption{\it Top-5 Categorical accuracy, stochastic model, plug-and-play approach}
    \end{subfigure}

      \end{tabular}

    \caption{\it Top-5 categorical accuracy of the subject recognition task as a function of $\alpha$ for both the deterministic and stochastic models. The baseline of simply using the pretrained FaderNet sampler is also shown for comparison. Note that the stochastic model is able to improve performance on the Top-5 categorical accuracy metric for some $\alpha$ values when compared to the baseline model.}
    \vspace{-.2in}
    \label{fig:subjectAccuracy}
\end{figure}

Figures \ref{fig:pgenderAdv}, \ref{fig:pgenderPAP} and \ref{fig:subjectAccuracy} show more directly interpretable results on privacy and utility performance as a function of $\alpha$. Figures \ref{fig:pgenderAdv} and \ref{fig:pgenderPAP} shows how the output probabilities of the gender classification model approach the prior distribution of the dataset as $\alpha$ increases for the adversarial and plug-and-play approaches respectively. We see that images from both female and male subjects produce output gender probabilities close to the dataset prior even for relatively low $\alpha$ values. Figure \ref{fig:subjectAccuracy} shows how the top-5 categorical accuracy of the subject recognition task varies across different $\alpha$ values and training modalities.
These results suggest that under these conditions we can achieve almost perfect privacy while maintaining reasonable utility performance.

{
We also show that retraining the privacy task adversarially with the sanitization function does not sacrifice performance on unfiltered data. Figure \ref{fig:pgenderRaw} in the supplementary material section shows how the output probabilities of the gender classification model on unfiltered data after adversarial training either maintains or improves performance when compared with the original gender classification model. This means that if the user decides not to block the gender (private data), the same system maintains its performance (and so does the utility/identity).
} 

We show the loss evolution of the sanitization task (Eq.\ref{eq:sanitizationLoss}) and the privacy task (Eq.\ref{eq:privatizationLoss}) across batch iterations for different $\alpha$ values in figures \ref{fig:lossesDet} and \ref{fig:lossesSto} in the supplementary material section. We observe that all losses settle quickly, and that overall the privacy loss grows over time as it is unable to infer the sensitive variable on the sanitized data.

\noindent
\textbf{{An Onboard Implementation-}}
The framework here proposed calls for the user to decide what to share and what not to share, and as such it renders itself for the sanitization function to be implemented as close as possible to the data acquisition, clearly before the user sharing her/his data with the service provider.
To illustrate this concept, we have designed a hardware prototype, described in the supplementary information. In brief, the system consists of an end device, capturing user visual data, and an entity, processing the user data. The end device  captures visual data, performs face detection, applies a privacy filter (per the user privacy requirements), and conveys the sanitized data to the service provider. The entity applies the algorithms associated with the utility and privacy tasks to the sanitized data. The respective results are displayed in a user interface.
We tested the performance of the trained sanitation functions over images acquired with the proposed hardware implementation. Four new subjects provided $104$ training images total, the subject identification network was retrained to also recognize these new subjects, in addition to all the subjects already present in the FaceScrub database.
The sanitization functions were not retrained on the newly acquired images. Figure \ref{fig:hwResults} in the supplementary material shows representative results on the stochastic plug-and-play sanitization function over $20$ test images. The sanitization function preserved utility performance while successfully blocking inference on the privacy variable.

\section{Connections with the Literature}
\label{sec:literature}

Our proposed framework and perspectives are new but these also connect to various elements in the privacy, statistics, and machine learning literature. Our goal here is to describe how our concepts relate to and depart from key concepts in this vast literature.

\noindent
\textbf{{Differential Privacy-}} Differential privacy and (local) differential privacy based data release mechanisms are often considered to be the gold standard in privacy \cite{Dwork2008}. Data release mechanisms based on (local) differential privacy -- which are also based on the application of some form of sanitization function to the original data -- ensure it will be almost impossible for a data receiver to distinguish whether or not there is a certain element in a dataset/database or a certain attribute in the data held by the data holder. However, such data release mechanisms can also result in substantial utility loss \cite{Dwork2010, Hardt2016, Kifer2011a,  Kifer2014}.
Our user-specific privacy collaborative approach departs from differential privacy in various ways. In particular, by building upon the premise that it is of interest both for users and service providers to collaborate to achieve privacy (and utility), we design data release mechanisms -- our privacy filter -- that achieve both utility and privacy, that are user-specific, and that are transparent to the service provider existing algorithms/architectures.

\noindent
\textbf{{Information Bottleneck and Privacy Funnel-}} The privacy funnel -- a concept closely related to the information bottleneck \cite{IB} -- has also been recently considered as a basis for the design of data release mechanisms \cite{funnel} (see also \cite{Wang2018}). In particular, this approach can lead to data release mechanisms that strike a balance between the level of utility and the level of privacy.
Both our approach and the privacy funnel (and its variants) can lead to data release mechanisms that meet specific user privacy requirements. However, it is not entirely clear how to design 'plug-and-play' privacy filters using the privacy funnel without the collaborative framework proposed here.

\noindent
\textbf{{Adversarial Networks and Adversarial Examples-}}
Adversarial game-type models as the one here presented clearly remind us of the type of works represented by generative adversarial networks (GANs) \cite{Goodfellow2014}. GANs are designed to learn the data (feature) distribution, while in our framework the task is to preserve privacy while guaranteeing utility, maintaining only the utility-needed data structure. As such, the proposed model addresses two goals, while GANs have the single goal of distribution learning (implemented though via the game-theory approach of two competing terms, not to confuse with two competing explicit tasks). We use the same general concept in \cite{Goodfellow2014} of training for adversarial goals, as also exploited in some of the works described below. Similarly, the literature on adversarial examples, \cite{Adversarial}, connects to our work, while at the same time being very different. Our goal is to preserve utility while creating uncertainty in the private data, while adversarial examples do not guarantee utility or privacy, but rather are targeted to make the system fail (note that making the system always fail in gender for example will actually reveal the gender with full certainty).

\noindent
\textbf{{Removing Nuisance/Attribute Variable-}}
A series of works have shown how to produce signals that preserve certain properties while removing others \cite{FaderNetworks,Pivot,ControllableInvariance}. While these works share certain architectural similarities with ours, and can be used as components or complementary to the framework here proposed, they are different both in goals and accomplishments. Works such as \cite{FaderNetworks} produce images without the nuisance, without guaranteeing neither the utility nor the privacy, e.g., producing a female version of a male face does not guarantee the utility (recognition) or even the privacy (gender). This is demonstrated in Section \ref{sec:results}. The use of GANs in \cite{Pivot} is in order to address robustness of the classifier/regressor to nuisance variables. Same goal motivates \cite{ControllableInvariance}, which contrary to us trains both the utility and the privacy classifiers, and both can only handle filtered data (the training is done into arbitrary spaces). The possibility to handle both filtered and unfiltered data and to not retrain all system components is critical for the utility provider adoption of a framework as the one here developed. 
Note also that the image example in \cite{ControllableInvariance} uses privacy and utility tasks that are statistically independent, whereas the pair of tasks we chose are very strongly related (there is an injective mapping from subjects to gender), thereby being a much harder problem.
Finally, we should make a clear distinction between aiming at making sure the output of the predictor/classifier is independent of the nuisance variable like in these works, which is more related to the closer topic of fairness (one of the examples addressed
in \cite{ControllableInvariance}), and the aim of privacy, which needs to make sure that the nuisance/sensitive variable is independent of the encoded features. 

\noindent
\textbf{{Protecting Training Data-}}
The remarkable success of machine learning
has led to intense interest in the privacy of the data used for training, since the released training models
contain information about it, e.g., \cite{Carlini2018,Hitaj2017}. Efforts to protect such data have been made, in particular, combining tools from differential privacy with parallel and independent training using data partitions and teacher/student paradigms \cite{Acs2017,PATE}. Contrary to these works, we are not concerned here with the protection of the training data but some of the attributes of the testing data, which depending on the users desires about whether these should or should not be revealed. In spite of these fundamental differences, these efforts do provide some insights for potential future directions for the challenge here addressed, in particular the use of different combinations and partitions of data for training the utility and privacy components, including potentially
training them with different combinations of original and filtered data and of true and random labels. Studying the utility training with the filtered data produced by our algorithm is a subject of interest and future work (\cite{Carlini2018} investigated training with sanitized data in their framework). The same applies to the important subject of fairness, and investigating the possibility of training with data that preserves utility but blocks potential biasing information, as produced by our filter, is of value as well (see also \cite{ControllableInvariance} for a contribution in this direction).

\section{Concluding Remarks}
\label{sec:conclusion}

We introduced a new paradigm where users and entities collaborate to achieve both utility and privacy per a user specific requirements. One salient feature of this paradigm is that it can be completely transparent -- involving only the use of a simple user-specific privacy filter applied to user data -- in the sense that it requires otherwise no modifications to the system infrastructure, including the service provider algorithmic capability, in order to achieve both utility and privacy.

Representative architectural and system concepts, and results, suggest that such a collaborative based user-controlled privacy approach can be achieved. While the results here presented clearly show the potential of this approach, much has yet to be done, from theoretical foundations to algorithm development to practical architectures, and extensions to scenarios involving multiple users with different utility and privacy requirements and scenarios involving other data types.



\section*{Acknowledgments}
Work partially supported by NSF, ONR, NGA, and ARO.

\bibliographystyle{ieee}
\bibliography{library_NIPS.bib}

\newpage

\begin{center}
\textbf{{\Large Supplementary Material}}
\end{center}

\section*{Performance of Gender Classification Model on Unfiltered Data}

\begin{figure}[!ht]
    \centering
    
    \begin{tabular}{cc}
    \begin{subfigure}[b]{0.4\textwidth}
        \includegraphics[width=\textwidth]{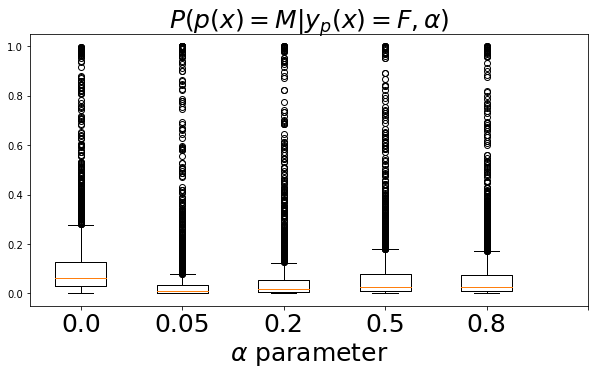}
        \caption{\it $P(p(x) = M | y_p(x) = F, \alpha)$, Deterministic model}
    \end{subfigure}&
    
        \begin{subfigure}[b]{0.4\textwidth}
        \includegraphics[width=\textwidth]{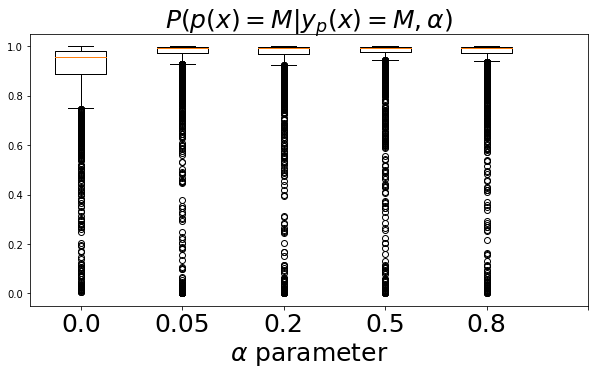}
        \caption{\it $P(p(x) = M | y_p(x) = M, \alpha)$, Deterministic model}
    \end{subfigure}\\
    
    ~ 

    \begin{subfigure}[b]{0.4\textwidth}
        \includegraphics[width=\textwidth]{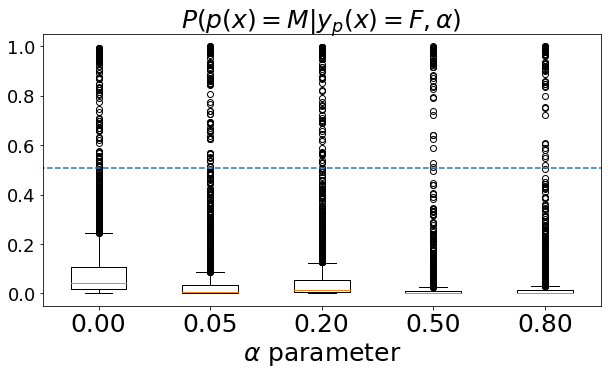}
        \caption{\it $P(p(x) = M | y_p(x) = F, \alpha)$, Stochastic model}
    \end{subfigure}&
    
        \begin{subfigure}[b]{0.4\textwidth}
        \includegraphics[width=\textwidth]{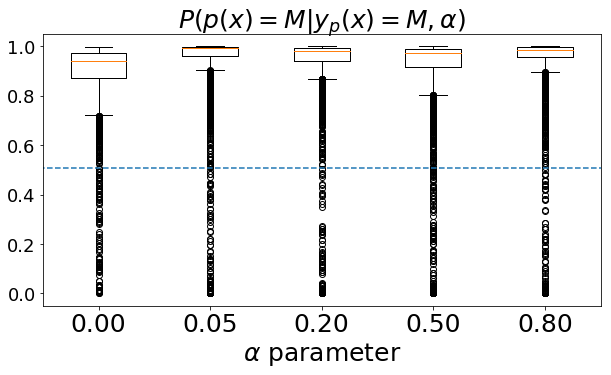}
        \caption{\it $P(p(x) = M | y_p(x) = M, \alpha)$, Stochastic model}
    \end{subfigure}
    
    ~ 
      \end{tabular}

    \caption{\it Detailed breakdown of the gender probabilities computed by the adversarially trained privacy task as a function of $\alpha$ for the test dataset over unfiltered images. Whisker plots show median and interquartile ranges, with outliers shown as circles.
    The first column shows the output probability of being male (M) as computed by the gender discriminator over all female (F) subjects. Second column shows the same output probabilities for all male subjects. Results are shown as a function $\alpha$, where $\alpha=0$ is the performance of the untrained gender detector over raw data.}
    \label{fig:pgenderRaw}
\end{figure}

We show how the adversarially trained privacy task maintains or improves performance on unfiltered data on Figure \ref{fig:pgenderRaw}. This empirically shows that there is little to no downside to this adversarial retraining of the privacy function.

\FloatBarrier

\section*{Details on Evolution of Training Losses Across Iterations}

\begin{figure}[!ht]
    \centering
    
    \begin{tabular}{cc}
    \begin{subfigure}[b]{0.4\textwidth}
        \includegraphics[width=\textwidth]{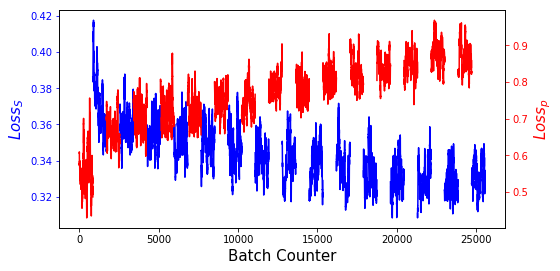}
        \caption{$\alpha$ = 0.05}
    \end{subfigure}&
    
        \begin{subfigure}[b]{0.4\textwidth}
        \includegraphics[width=\textwidth]{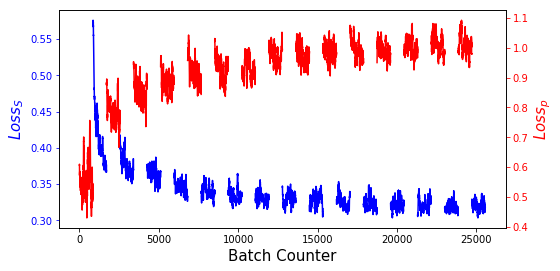}
        \caption{$\alpha$ = 0.2}
    \end{subfigure}
    
    ~ 
      \end{tabular}
      
         \begin{tabular}{cc}
    \begin{subfigure}[b]{0.4\textwidth}
        \includegraphics[width=\textwidth]{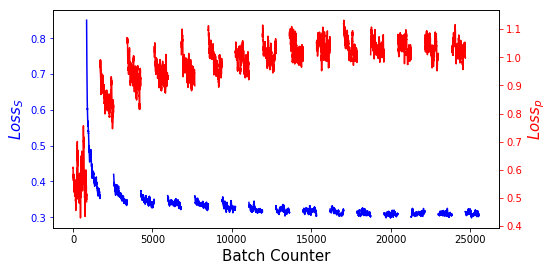}
        \caption{$\alpha$ = 0.5}
    \end{subfigure}&
    
        \begin{subfigure}[b]{0.4\textwidth}
        \includegraphics[width=\textwidth]{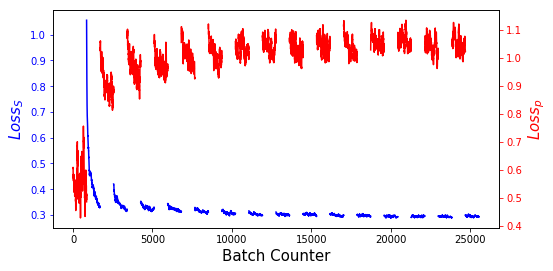}
        \caption{$\alpha$ = 0.8}
    \end{subfigure}
    
    ~ 
      \end{tabular}
    \caption{\it Evolution of the privacy loss ($Loss_p$) and sanitization loss ($Loss_S$) as a function of training time.Results are shown on the deterministic model trained adversarially. We observe that the both losses settle quickly, and that overall, the privacy loss grows over time as it is unable to infer the sensitive variable on the sanitized data.}
    \label{fig:lossesDet}

\end{figure}

\begin{figure}[!ht]
    \centering
    
    \begin{tabular}{cc}
    \begin{subfigure}[b]{0.4\textwidth}
        \includegraphics[width=\textwidth]{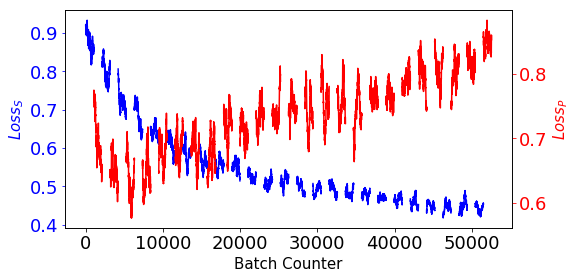}
        \caption{$\alpha$ = 0.05}
    \end{subfigure}&
    
        \begin{subfigure}[b]{0.4\textwidth}
        \includegraphics[width=\textwidth]{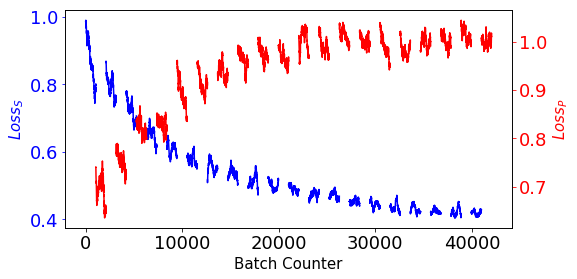}
        \caption{$\alpha$ = 0.2}
    \end{subfigure}
    
    ~ 
      \end{tabular}
      
         \begin{tabular}{cc}
    \begin{subfigure}[b]{0.4\textwidth}
        \includegraphics[width=\textwidth]{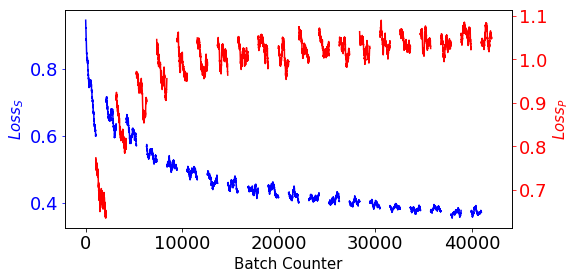}
        \caption{$\alpha$ = 0.5}
    \end{subfigure}&
    
        \begin{subfigure}[b]{0.4\textwidth}
        \includegraphics[width=\textwidth]{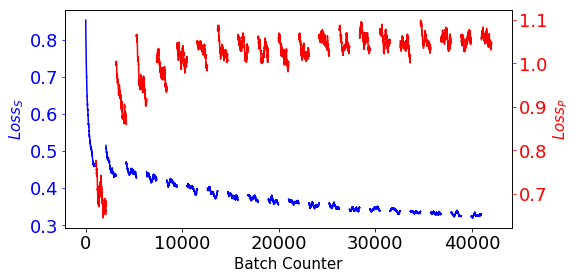}
        \caption{$\alpha$ = 0.8}
    \end{subfigure}
    
    ~ 
      \end{tabular}
    \caption{\it Evolution of the privacy loss ($Loss_p$) and sanitization loss ($Loss_S$) as a function of training time. Results are shown on the stochastic model trained adversarially. We observe that both losses settle quickly, and that overall, the privacy loss grows over time as it is unable to infer the sensitive variable on the sanitized data.}
    \label{fig:lossesSto}

\end{figure}

\begin{figure}[!ht]
    \centering
    
    \begin{tabular}{cc}
    \begin{subfigure}[b]{0.4\textwidth}
        \includegraphics[width=\textwidth]{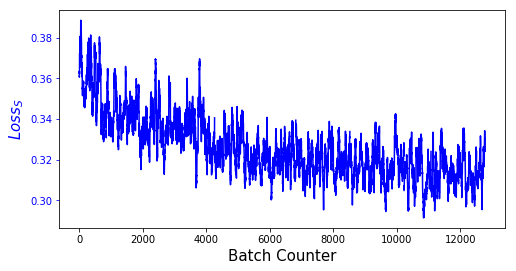}
        \caption{$\alpha$ = 0.05}
    \end{subfigure}&
    
        \begin{subfigure}[b]{0.4\textwidth}
        \includegraphics[width=\textwidth]{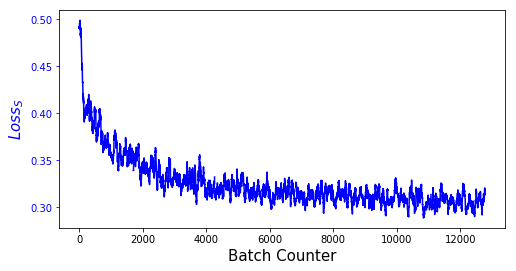}
        \caption{$\alpha$ = 0.2}
    \end{subfigure}
    
    ~ 
      \end{tabular}
      
         \begin{tabular}{cc}
    \begin{subfigure}[b]{0.4\textwidth}
        \includegraphics[width=\textwidth]{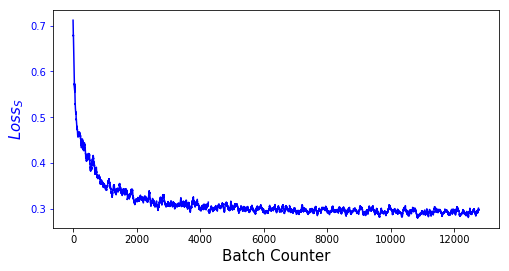}
        \caption{$\alpha$ = 0.5}
    \end{subfigure}&
    
        \begin{subfigure}[b]{0.4\textwidth}
        \includegraphics[width=\textwidth]{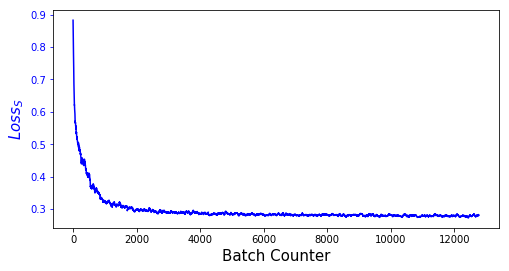}
        \caption{$\alpha$ = 0.8}
    \end{subfigure}
    
    ~ 
      \end{tabular}
    \caption{\it Evolution of the sanitization loss ($Loss_S$) as a function of training time. Results are shown on the deterministic model trained using the plug-and-play approach. The loss quickly settles to its final value.}
    \label{fig:lossesDetPap}

\end{figure}

\begin{figure}[!ht]
    \centering
    
    \begin{tabular}{cc}
    \begin{subfigure}[b]{0.4\textwidth}
        \includegraphics[width=\textwidth]{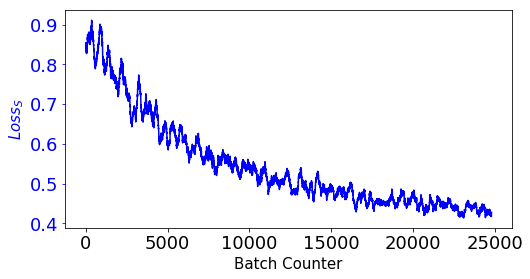}
        \caption{$\alpha$ = 0.05}
    \end{subfigure}&
    
        \begin{subfigure}[b]{0.4\textwidth}
        \includegraphics[width=\textwidth]{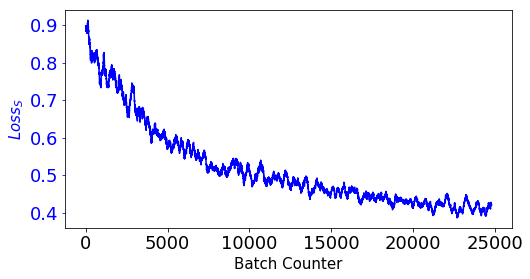}
        \caption{$\alpha$ = 0.2}
    \end{subfigure}
    
    ~ 
      \end{tabular}
      
         \begin{tabular}{cc}
    \begin{subfigure}[b]{0.4\textwidth}
        \includegraphics[width=\textwidth]{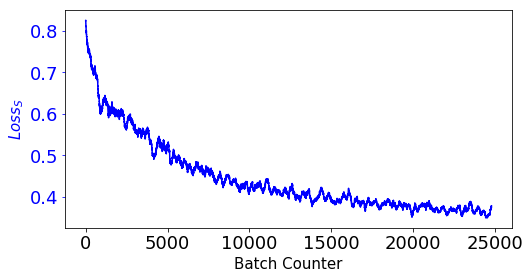}
        \caption{$\alpha$ = 0.5}
    \end{subfigure}&
    
        \begin{subfigure}[b]{0.4\textwidth}
        \includegraphics[width=\textwidth]{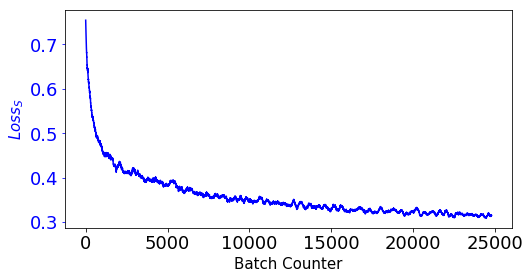}
        \caption{$\alpha$ = 0.8}
    \end{subfigure}
    
    ~ 
      \end{tabular}
    \caption{\it Evolution of the sanitization loss ($Loss_S$) as a function of training time. Results are shown on the stochastic model trained using the plug-and-play approach. The loss quickly settles to its final value.}
    \label{fig:lossesStoPap}

\end{figure}

Here we show the loss evolution of the sanitization task ($Loss_S$) when using the plug-and-play and adversarial training for both the deterministic (figures \ref{fig:lossesDet} and \ref{fig:lossesDetPap}) and stochastic architectures  (figures \ref{fig:lossesSto} and \ref{fig:lossesStoPap}). The privacy loss ($Loss_P$) is also shown on the adversarial training approach. The privacy and sanitization tasks are trained alternatively every epoch. Overall, $Loss_P$ grows over time as it is unable to infer the sensitive variable on the sanitized data.

\FloatBarrier
\subsection*{Details on the Onboard Implementation}

\begin{figure}[!ht]
\centering
\includegraphics[width=0.8\columnwidth]{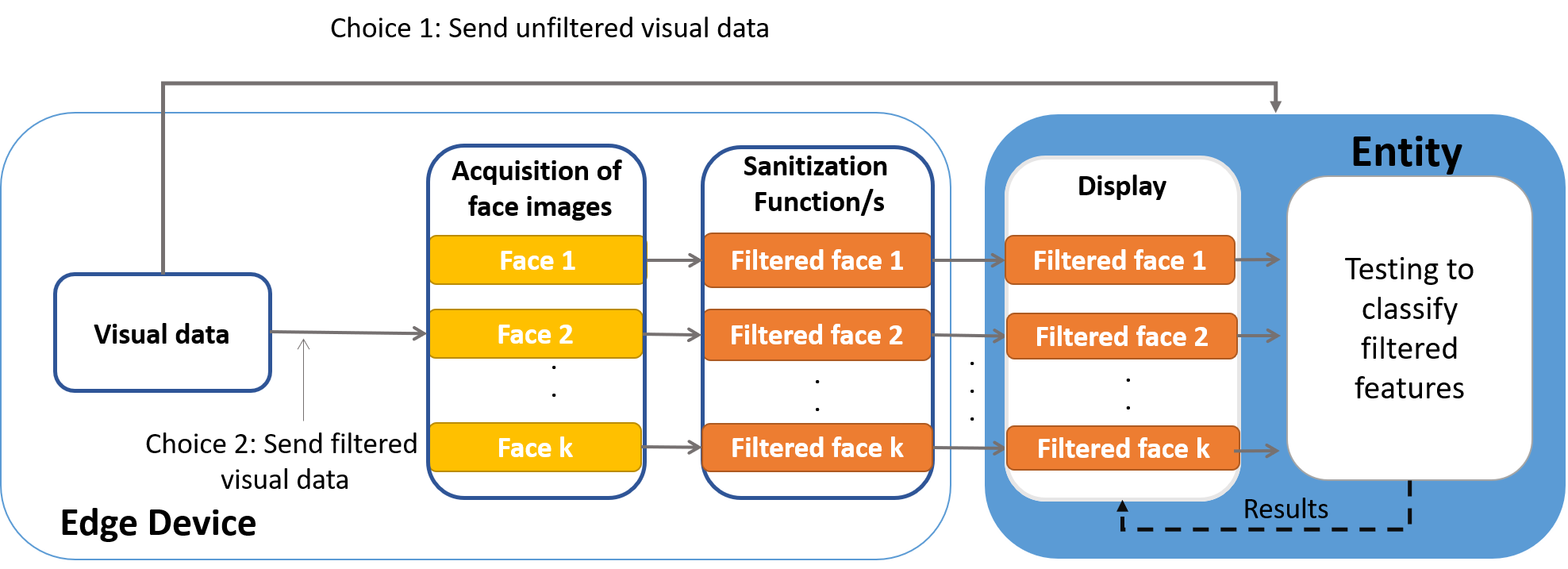}
\caption{\it Edge device and entity tasks.}
\label{fig:edgedevice}
\end{figure}

This section describes a proof-of-concept prototype of our proposed collaborative privacy based system. The main components of this system are an edge device and an entity. The edge device is responsible to obtain visual user data, pre-process the visual data, and convey it to the entity. The pre-processing phase occurs when the user chooses to protect data that she/he does not want the entity to infer. This phase includes several tasks:
\begin{enumerate}
    
    \item The first task is to locate a face (or multiple faces) within the acquired visual data. In particular, we use the face detector from dlib library \cite{dlib09}, consisting of a combination of the Histogram of Oriented Gradients (HOG) feature with a linear classifier, an image pyramid, and sliding window detection scheme. After this, the found faces are treated as individual images.  
    
    \item The second task is to filter what the user would like the entity not to be able to infer. Each detected face is filtered according to the learned sanitization function responsible for the selected feature the user wishes to protect. This process transforms the data onto another piece of data, in the same space, such that utility is preserved but privacy is protected.
    
    \item Finally, the data is conveyed to the entity using TCP/IP protocols.  
    
\end{enumerate}
Figure \ref{fig:edgedevice} summarizes these various steps. 

\begin{figure}[!ht]
\centering
\includegraphics[width=0.8\columnwidth]{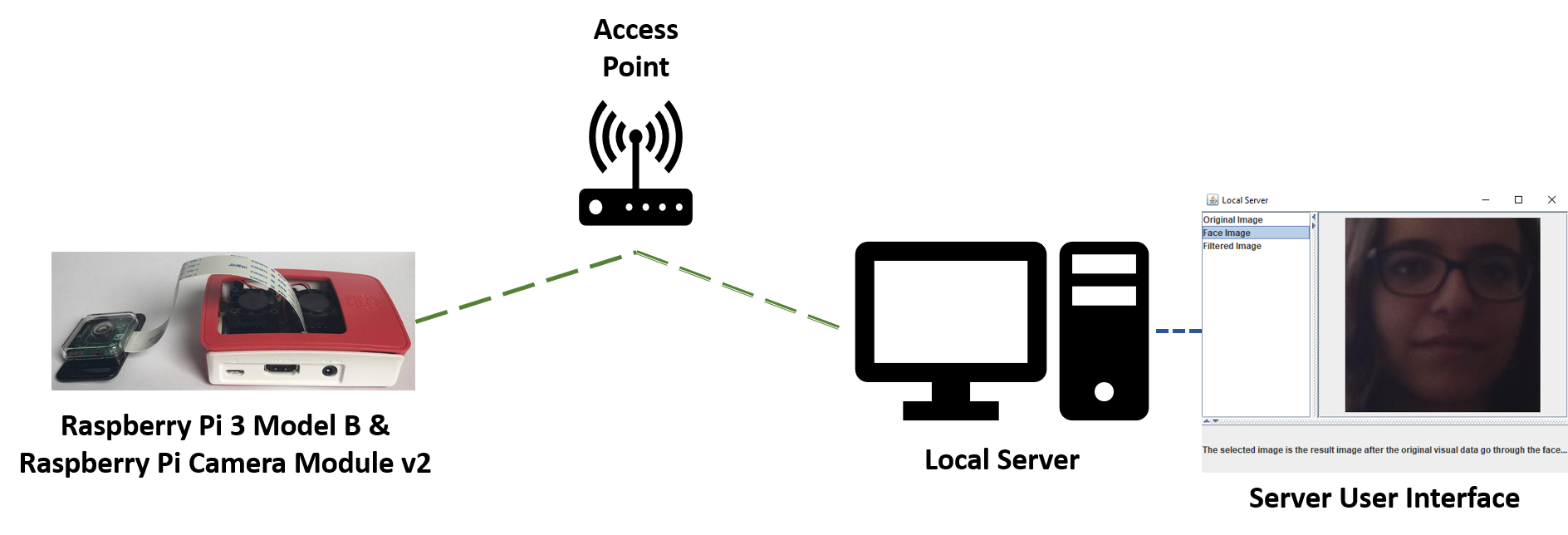}
\caption{System overview.}
\label{fig:systemoverview}
\end{figure}

A Raspberry Pi 3 Model B board \cite{Rpi3} with quad-core 1.2GHz ARM Cortex A53 (ARMv8) CPU and 1GB LPDDR2 RAM running at 900MHz was used as a system’s edge device. The operating system installed on the board is Raspbian and it is powered by a 5.1 Volts micro USB supply. In order to obtain the visual data, a Raspberry Pi camera module v2 \cite{Rpicamera} was attached to the Raspberry Pi board via a ribbon cable. The camera module v2 is based on the 8-Megapixels resolution Sony IMX219 sensor. The images are sent using the embedded WIFI module of the Raspberry Pi 3. 
See Figure \ref{fig:systemoverview} for these components.

For the purposes of the entity component, we developed a local server. This server communicates with the Raspberry Pi 3 using a Wireless Local Area Network (WLAN) to receive the user data. There, the received information is saved and displayed on a user interface. In parallel, a number of algorithms are used in order to try to infer information from the visual data. The classification results are displayed in the same interface as well.

\begin{figure}[!ht]
    \centering
    
    \begin{tabular}{ccc}
    \begin{subfigure}[b]{0.3\textwidth}
        \includegraphics[width=\textwidth]{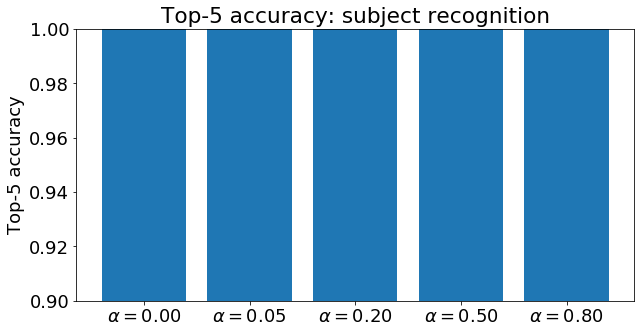}
        \caption{\it Top-5 categorical accuracy, stochastic model.}
    \end{subfigure}&
    
    \begin{subfigure}[b]{0.3\textwidth}
        \includegraphics[width=\textwidth]{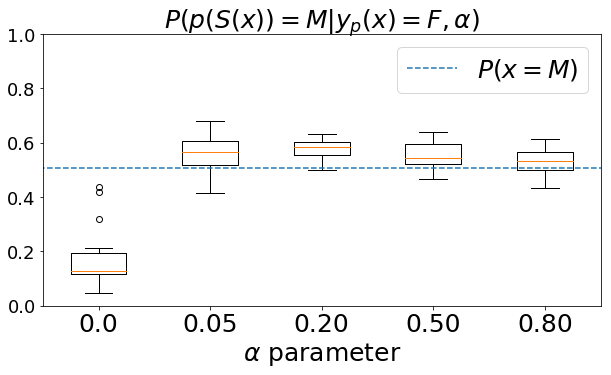}
        \caption{\it $P(p(S(x)) = M | y_p(x) = F, \alpha)$, Stochastic model}
    \end{subfigure}&
    
        \begin{subfigure}[b]{0.3\textwidth}
        \includegraphics[width=\textwidth]{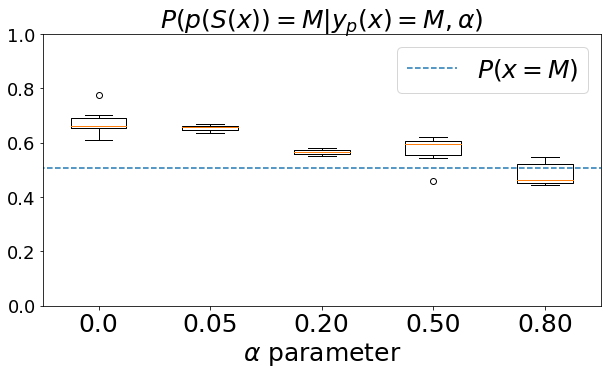}
        \caption{\it $P(p(S(x)) = M | y_p(x) = M, \alpha)$, Stochastic model}
    \end{subfigure}
      \end{tabular}

    \caption{\it Detailed breakdown of the top-5 categorical accuracy and gender probabilities computed by the utility and privacy tasks as a function of $\alpha$ on data from the hardware-acquired dataset. The sanitization function shown here is the stochastic model trained using the plug-and-play approach.}
    \label{fig:hwResults}
\end{figure}

Figure \ref{fig:hwResults} shows representative results on the stochastic plug-and-play sanitization function over $20$ test images. The sanitization function preserved utility performance while successfully blocking inference on the privacy variable.

\end{document}